\documentclass[runningheads]{llncs}

\usepackage[mobile]{eccv}

\definecolor{cvprblue}{rgb}{0.21,0.49,0.74}
\definecolor{srcblue}{RGB}{0,150,255}
\definecolor{drvorange}{RGB}{255,170,50}

\usepackage{multirow}
\usepackage{pifont}

\makeatletter
\renewcommand{\paragraph}{%
  \@startsection{paragraph}{4}%
  {\z@}{0.2ex \@plus 0.3ex \@minus .2ex}{-1em}%
  {\normalfont\normalsize\bfseries}%
}
\makeatother

\usepackage{siunitx}
\usepackage{eccvabbrv}

\usepackage{bm}
\usepackage{graphicx}
\usepackage{booktabs}
\usepackage{wrapfig}
\usepackage[table,xcdraw,dvipsnames]{xcolor}
\usepackage[accsupp]{axessibility}  %

\usepackage{hyperref}

\usepackage{orcidlink}

\author{
Kaiwen Jiang\inst{1}\thanks{This project was initiated and substantially carried out during an internship at NVIDIA.} \and
Xueting Li\inst{2} \and
Seonwook Park\inst{2} \and
Ravi Ramamoorthi\inst{1} \and
Shalini De Mello\inst{2} \and
Koki Nagano\inst{2}
}

\institute{
University of California, San Diego, United States
\and
NVIDIA, United States
}

\authorrunning{Jiang et al.}

\begin{document}

\title{Instant Expressive Gaussian Head Avatars at Over 100 FPS}

\maketitle

\begin{figure}[t]
    \centering
    \includegraphics[width=\linewidth]{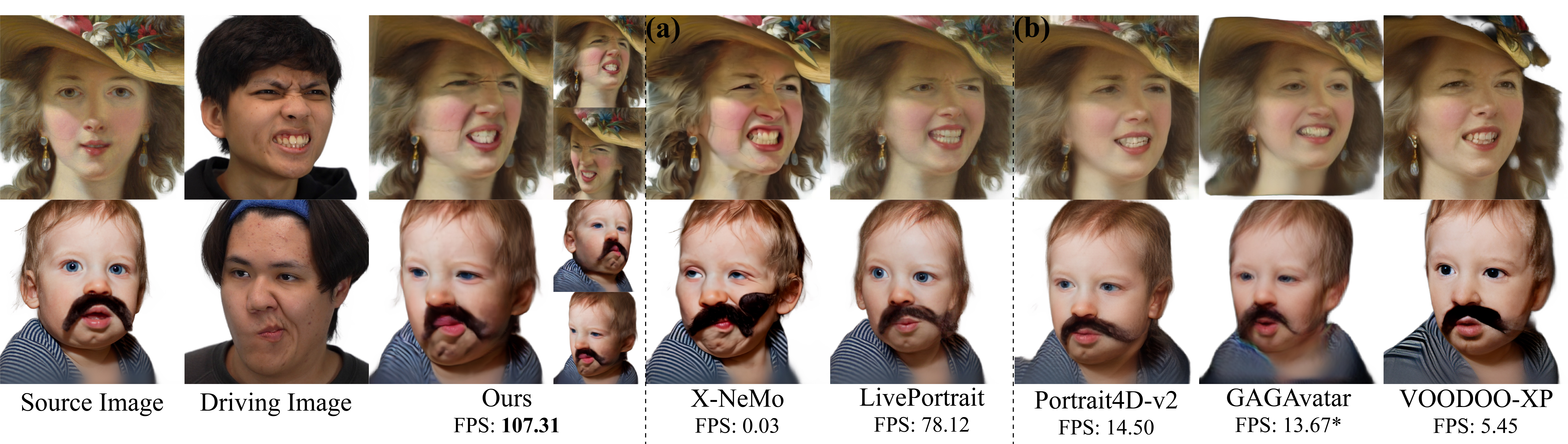}
    \caption{We present an instant {feedforward} encoder that transforms an in-the-wild source image into an animatable 3D avatar. Our method introduces a fast, consistent yet expressive 3D animation representation. 
    Given a driving image, we evaluate both the expression transfer quality and the animation speed (measured as ``FPS'' on an NVIDIA 6000 Ada GPU) against \textbf{(a)} 2D diffusion- or GAN-based methods and \textbf{(b)} 3D-aware methods. 
    In the first row, Portrait4D-v2 \cite{deng2024portrait4dv2}, GAGAvatar \cite{gagavatar} and VOODOO-XP \cite{voodooxp} fail to faithfully transfer expressions, particularly around the nasal wrinkles. LivePortrait \cite{guo2024liveportrait} is inaccurate at eyes. 
    In the second row, the baby wears a fake mustache as a decoration. X-NeMo distorts identity and adds a hallucinated mustache. Other methods cannot deal with the asymmetric expression in the driving image well.
    In contrast, ours not only accurately transfers expressions but also achieves high animation speed and consistent pose control. Insets show our rendered results under different poses.
    \emph{FPS is measured in a fully end-to-end streaming setting throughout the paper, where incoming frames are processed sequentially and all modules, including preprocessing and fitting steps, are accounted for across all baselines.}
    }
    \label{fig:teaser}
\end{figure}

\begin{abstract}
Portrait animation has witnessed tremendous quality improvements thanks to recent advances in video diffusion models. However, these 2D methods often compromise 3D consistency and speed, limiting their applicability in real-world scenarios, such as digital twins or telepresence. In contrast, 3D-aware feedforward facial animation methods -- built upon 3D representations, such as neural radiance fields or Gaussian splatting -- ensure 3D consistency and achieve faster inference speed, but come with inferior expression details. In this paper, we address this \textit{portrait animation trilemma} (speed, 3D consistency, and expressiveness) and propose a pipeline that instantly converts an in-the-wild single image into a 3D-consistent, fast yet expressive animatable representation via a feed-forward encoder. 
Unlike previous computationally intensive global fusion mechanisms (e.g., multiple attention layers) for fusing 3D structural and animation information, our design employs an efficient lightweight local fusion strategy to achieve high animation expressivity. Furthermore, our animation representation is decoupled from the face's 3D representation and learns motion implicitly from data, eliminating the dependency on pre-defined parametric models that often constrain animation capabilities. Our method runs at 107.31 FPS for animation and pose control, representing a 3-4 order of magnitude speedup versus the state of the art while achieving comparable animation quality, thus surpassing alternative designs that trade speed for quality or vice versa. Project Page: \url{https://research.nvidia.com/labs/amri/projects/instant4d}
\end{abstract}
    
\section{Introduction}
\label{sec:intro}
{Creating a 4D digital twin from a single facial image that supports both 3D viewpoint control and animation} is a long-standing goal in computer vision and graphics. Real-time animatable %
synthesis and animation control of photorealistic digital humans is essential for developing AR/VR, video conferencing, and agentic AI applications.

\begin{wrapfigure}{r}{0.5\columnwidth}
    \centering
    \includegraphics[width=\linewidth]{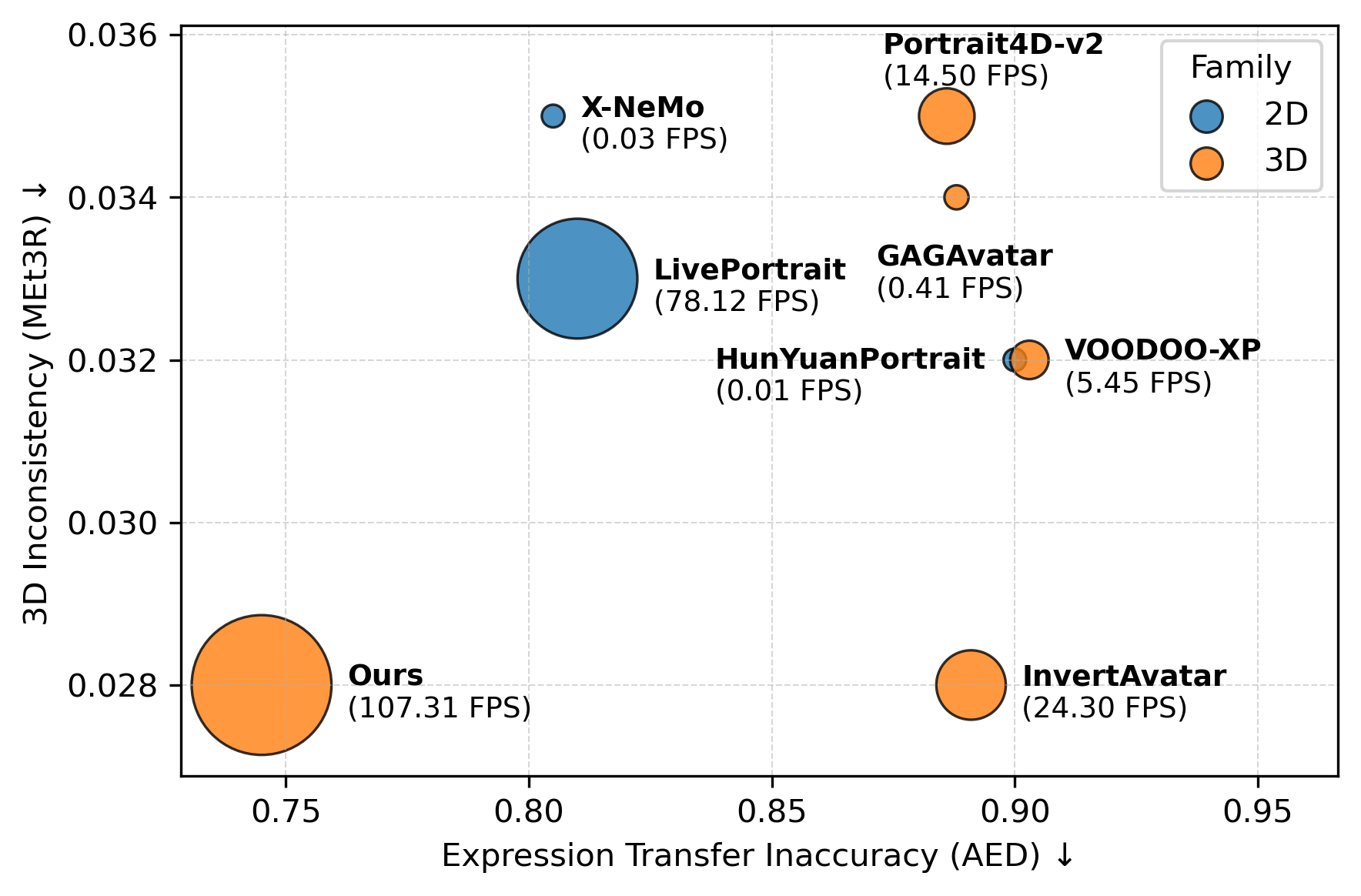}
    \caption{A portrait animation trilemma: existing methods do not simultaneously achieve 3D consistency, speed, and expressive expression transfer. 
    We provide a visualization of the quantitative comparison in terms of 3D inconsistency (measured by MEt3R $\downarrow$), expression transfer inaccuracy (measured by AED $\downarrow$) and animation speed (measured by FPS $\uparrow$, visualized as the size of the circle) with other 2D- or 3D-based baselines, including \cite{zhao2024invertavatar,deng2024portrait4dv2,voodooxp,xu2025hunyuanportrait,gagavatar,guo2024liveportrait,zhao2025x}, using the task of cross-reenactment. 2D methods tend to appear on the upper left (better expression transfer accuracy; worse 3D consistency) while 3D methods tend to appear on the lower right (worse expression transfer accuracy; better 3D consistency). Our method is \textbf{3-4} orders of magnitude faster than diffusion based models~\cite{xu2025hunyuanportrait,zhao2025x} while simultaneously achieving better 3D consistency and expression transfer accuracy, appearing on the lower left. %
    }
    \vspace{2em}
    \label{fig:method-comparison}
\end{wrapfigure}

Achieving such comprehensive 4D control has been historically challenging. With the advent of radiance fields, including neural radiance fields (NeRFs) \cite{mildenhall2020nerf} and 3D Gaussians \cite{3dgs}, previous 3D-aware face animation works~\cite{li2024generalizable,chu2024gpavatar,zhao2024invertavatar,gagavatar,sun2023next3d} have achieved real-time %
and consistent animation with photo-realistic view synthesis by using parametric models \cite{blanz1999morphable,3dmm,FLAME:SiggraphAsia2017}. However, parametric models inherently limit the animation {quality}. Follow-up methods \cite{kirschstein2025avat3r,voodooxp,deng2024portrait4dv2,deng2023portrait4d,voodoo3d} therefore resort to learning the animation purely from data. Nevertheless, their representations entangle 3D structure and animation (e.g., {via} global residual triplanes \cite{eg3d} or feature maps), requiring computationally expensive attention mechanisms to repeatedly fuse 3D structure and motion at every animation step.
Meanwhile, the introduction of 2D diffusion models~\cite{xu2025hunyuanportrait,zhao2025x,cui2025hallo3,xie2024x,ma2024followyouremoji,qiu2025skyreels,guo2024liveportrait,cao2025uni3c,cheng2025wan,sang2025lynx,ren2025gen3c,kuang2024cvd} into portrait animation has brought the expressivity of achieved facial animation to a whole new level~\cite{zhao2025x, xu2025hunyuanportrait}. However, these methods often suffer from 3D inconsistency and remain slow due to the expensive denoising process, preventing them from being used in a real-time system. Fig.~\ref{fig:method-comparison} shows a quantitative comparison where existing methods fail to excel at \emph{all three criteria}: speed, 3D consistency and expression transfer {expressivity}. We call this new challenge \emph{the portrait animation trilemma}. 

There are existing methods (e.g., \cite{taubner2025cap4d,giebenhain2024npga,tang2025gaf,aneja2025scaffoldavatar, taubner2025mvp4d}) proposing to optimize a person-specific animatable 3D avatar from captured videos or generated images that achieve these three goals.
However, we seek to design a 3D-aware animation framework that \emph{instantly} encodes a facial image into an animatable 4D avatar, {while supporting} fast, consistent and detailed animation. Our key insights for solving this problem are twofold. First, we observe that large-scale 2D diffusion models already encode rich facial dynamics. We reinterpret them as motion priors for 3D-aware animation and introduce a principled mechanism to leverage their guidance while avoiding their inherent 2D inconsistencies and hallucinations.
Second, we show that achieving both efficiency and 3D consistency without sacrificing expressiveness requires a novel animation representation, which we propose. %

Specifically, we build upon 3D Gaussians \cite{3dgs} that makes dynamic deformation more efficient than NeRFs’ volumetric fields to propose an expressive yet efficient animation representation. 
We first encode an input image into triplanes \cite{lp3d} as an intermediate representation, from which we sample feature vectors to decode the Gaussian attributes. For each Gaussian, we encode its motion information in an auxiliary vector which is analogous to learned personalized animation bases in traditional blendshapes animation. This auxiliary vector is then combined with the driving signal to deform each Gaussian {associated with} the existing 3D structure, updating the 3D representation while keeping the animation representation both efficient and decoupled from the underlying 3D representation.

While previous works~\cite{aligngaussians,liu2024dynamic,wu20244d,lin2025movies} deform the 3D Gaussians in spatial space to model motion, we find it unable to capture expressive facial details. Instead, we propose to deform the Gaussians individually in the high-dimensional feature vector space, rather than in 3D spatial space, which offers better expressivity and is capable of capturing asymmetric expressions and details such as shadow changes and wrinkles (Fig.~\ref{fig:teaser}).

Typically, portrait animation methods, including ours, are trained with a self-reenactment objective using datasets that contain multiple expression-varying images per identity. %
Instead of training on real datasets, we construct a synthetic facial expression dataset via a state-of-the-art facial animation diffusion model \cite{zhao2025x} for transferring its expressive motion priors in a 3D consistent manner. To minimize the 3D inconsistency and hallucination issues of diffusion models, we first frontalize in-the-wild portraits and then synthesize expressions by the diffusion model only on frontal views, thereby avoiding hallucinations caused by viewpoint changes. Multi-view results are then separately synthesized during training using a pre-trained expert 2D-to-3D network, which enforces cross-view geometric consistency via multiview supervision. Therefore, our method is \emph{free of} inconsistency or hallucination issues. %
In summary, our main contributions include:
\begin{itemize}
    \item We design an expressive yet computationally-efficient animation representation for 3D Gaussians that achieves detailed animation for human faces (Sec.~\ref{sec:method-design}).
    \item We propose practical strategies to train this animation representation by leveraging guidance from existing diffusion-based methods (Sec.~\ref{sec:method-distill}). %
    \item Our method is the first to simultaneously achieve best 3D consistency, fast inference speed and detailed expressions such as wrinkles, and run orders of magnitude faster than {competing methods at the same level of expressivity.} (Fig.~\ref{fig:teaser},~\ref{fig:method-comparison}).
\end{itemize}

\section{Related Work}
\label{sec:related-work}
\paragraph{2D and 3D facial portrait animation.} 2D facial portrait animation
methods often feature a generative backbone (e.g., GAN \cite{goodfellow2014generative} or diffusion \cite{score_matching,ddpm}), which synthesizes driven faces given control signals. GAN-based methods~\cite{siarohin2019animating,siarohin2019first,ren2021pirenderer,wang2020one,mallya2022implicit,yin2022styleheat,gao2023high,doukas2021headgan, guo2024liveportrait,zhao2022thin,burkov2020neural,9878472,zhou2021pose,pang2023dpe,wang2022latent,10645735,wang2023progressive,drobyshev2022megaportraits,drobyshev2024emoportraits,xu2024vasa} feature fast inference using either explicit or implicit expression representations, but are limited by the capability of GAN models. A diffusion backbone--often pre-trained on large-scale internet data--provides much stronger synthesis capability, and has been {employed for} facial animation~\cite{liu2024towards,xu2024facechain,varanka2024towards,paskaleva2024unified, tian2024emo,zhao2025x,xie2024x,xu2025hunyuanportrait,qiu2025skyreels,wei2024aniportrait,xu2024hallo,cui2025hallo3,yang2024megactor,wang2024v,ma2024followyouremoji,chen2024echomimic}{, showing excellent expression transfer quality}. %
However, the repeated denoising steps trade speed for quality and are thus prohibitive for real-time applications. {They are also not 3D consistent.} %

Another line of methods builds explicit 3D representations for 3D talking heads, which improve 3D consistency. They typically rely on 3D morphable models (3DMM) \cite{blanz1999morphable, 3dmm, flame, giebenhain2024mononphm, he2025lam} or facial motion representations (e.g., rasterized coordinates~\cite{PNCC} or facial keypoints) as priors~\cite{tang20233dfaceshop, sun2023next3d, zhao2024invertavatar, wu2022anifacegan, xu2023omniavatar, li2024generalizable, export3d, hidenerf, giebenhain2024npga, ye2024real3d, siarohin2023unsupervised, gagavatar} {for animating the face}. 
However, morphable models inherently limit facial expressiveness, as their strong statistical priors, which are derived from a finite set of face scans and linear basis representations restrict motion to a narrow, predefined space. Therefore, another group of methods~\cite{voodoo3d, voodooxp, deng2024portrait4dv2, deng2023portrait4d} implicitly learns the motion as residual features to the triplanes \cite{eg3d} through data. Notably, Portrait4D \cite{deng2023portrait4d} transfers knowledge from synthetic data. However, in their animation representation, the global residual features coupled with dense attention mechanisms are computationally expensive to infer. We instead propose a local fusion mechanism that individually deforms each 3D Gaussian through a lightweight MLP based on a learned auxiliary vector that encodes all the motion information. %
\begin{figure}[t]
    \centering
    \includegraphics[width=0.95\linewidth]{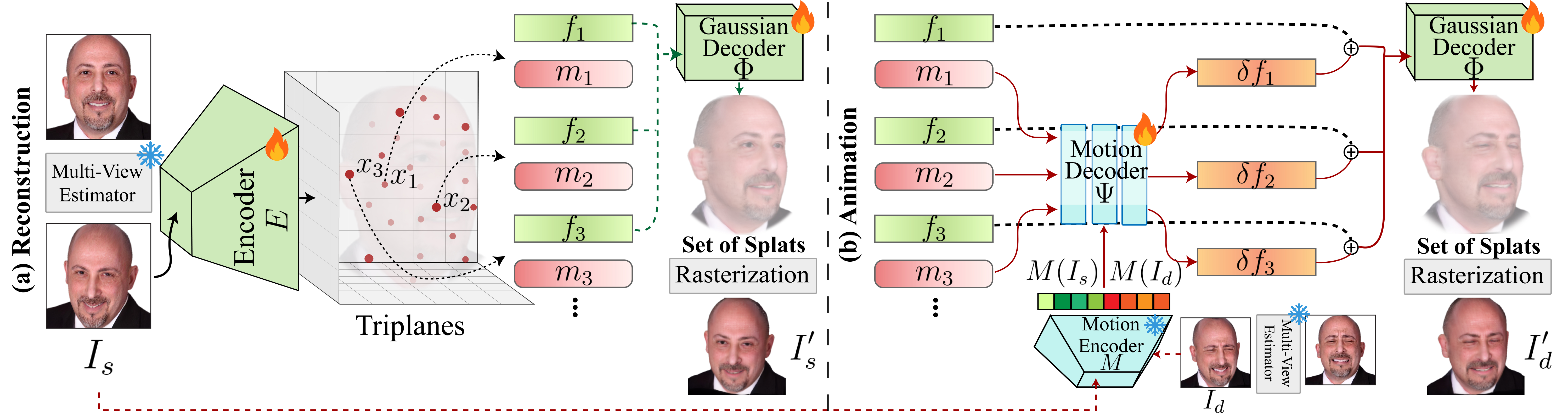}
    \caption{
    Overview of our \textbf{training} pipeline with the {two-part} self-reenactment task. \textbf{(a) Reconstruction}: Given a frontalized source frame with an expression synthesized by a pre-trained diffusion model~\cite{zhao2025x}, we first use a pre-trained multi-view estimator~\cite{lp3d} to generate {its another viewpoint} $I_s$. The encoder $E$ converts $I_s$ into triplanes, from which we sample feature vectors $f_1,f_2,\ldots$ and paired motion basis vectors $m_1,m_2,\ldots$. A Gaussian decoder $\Phi$ maps these features into a set of 3D Gaussians, forming a lifted 3D avatar for $I_s$, which we render at the viewpoint of $I_s$ as $I_s'$. \textbf{(b) Animation:} For the synthesized driving frame {of the same identity but with a different expression}, we similarly obtain its another viewpoint image %
    $I_d$. Both $I_s$ and $I_d$ are input into the motion encoder $M$ to produce motion coefficients $M(I_s)$ and $M(I_d)$. They are concatenated to condition a motion decoder $\Psi$ to predict residual features $\delta f_1, \delta f_2,\ldots$ from paired motion basis vectors. Adding these residuals to the original features and decoding them with $\Phi$ yields an animated set of Gaussians, which we render at the viewpoint of $I_d$ as $I_d'$.
    The loss is computed between $(I_s, I_s')$ and $(I_d, I_d')$. Fire icons denote trainable modules; snow icons denote frozen pre-trained modules.
    }
    \label{fig:pipeline}
\end{figure}

\paragraph{4D representation and rendering.}
Aiming at a native 4D representation, many works~\cite{park2021nerfies,park2021hypernerf,fang2022fast,Pumarola20arxiv_D_NeRF} build upon NeRFs to construct a spatial deformation field, which, however, is computationally expensive.
With the advent of Gaussian splatting \cite{3dgs}, another family of methods identify the potential of using the spatial deformation of 3D Gaussians to represent motion. One group of methods~\cite{wu20244d,aligngaussians,liu2024dynamic,giebenhain2024npga} learn an implicit representation, such as HexPlanes \cite{cao2023hexplane} or a coordinate-based MLP, to deform the Gaussians.
Another group of methods~\cite{zielonka2023drivable,giebenhain2024npga,yu25gaia,tang2025gaf, he2025lam} utilizes an explicit mesh to drive the 3D Gaussians. However, we propose to deform Gaussians in a high-dimensional feature space, thereby preserving greater expressivity (Sec.~\ref{sec:ablation}).

Notably, Avat3r \cite{kirschstein2025avat3r} and ScaffoldAvatar \cite{aneja2025scaffoldavatar} also use 3D Gaussians to build animatable avatars. However, they are not a feed-forward method from a single image and require either 3D GAN inversion or mesh tracking. 
Prior work \cite{taubner2025mvp4d,taubner2025cap4d} uses diffusion-generated multi-view images for animated 3D head synthesis, but relies on hours-long optimization to fit a single avatar. %
In contrast, we develop a generalizable {3D-consistent} framework that requires no tracking, supports instant encoding, and enables real-time {expressive} animation.

\section{Preliminaries}
\label{sec:method-preliminary}
\paragraph{3D avatar encoder and volume rendering.} We choose the state-of-the-art facial 2D-to-3D lifting encoder \cite{lp3d} as our architectural backbone for lifting a single-view image into a 3D avatar. Given a single image $I$, the encoder encodes it into triplanes $\{T_{xy},T_{yz},T_{zx}\}$~\cite{eg3d}, each of which is $\in\mathbb{R}^{256\times256\times32}$. These planes can then be used for rendering arbitrary viewpoints using volumetric rendering \cite{mildenhall2020nerf}. Specifically, for each queried 3D position $\mathbf{x}=(x,y,z)$, its corresponding feature vector ${f}(\mathbf{x})\in\mathbb{R}^{32}$ is retrieved by projecting $\mathbf{x}$ onto each of the three planes via bilinear interpolation and further aggregation{ across the three planes} by summation. %
A light-weight non-linear multi-layer perceptron (MLP) decoder then decodes the aggregated features into colors and densities for volume rendering. In this work, we extend this approach by adapting this NeRF-based encoder for encoding a single-view image into a set of \textbf{3D Gaussians} as explained later (Sec.~\ref{sec:method-design}).%

\paragraph{3D Gaussian splatting.} Kerbl et al.~\cite{3dgs} provides a differentiable and efficient solution to encoding a 3D scene as a set of anisotropic 3D Gaussian and rendering them efficiently into images. Specifically, each 3D Gaussian {is} parametrized by its position vector $\mathbf{\mu}\in\mathbb{R}^3$, scaling vector $\mathbf{s}\in\mathbb{R}^3$, quaternion vector $\mathbf{q}\in\mathbb{R}^4$, opacity $o\in\mathbb{R}$ and color $\mathbf{c}\in\mathbb{R}^3$. The final rendering color of a pixel is calculated by alpha-blending all 3D Gaussians overlapping the pixel. In the following discussion, we use the sub-script $i$ to denote that these quantities belong to the $i^\text{th}$ 3D Gaussian.

\section{Method}
\label{sec:method}
\paragraph{Overview.} Our overall training pipeline is shown in Fig.~\ref{fig:pipeline}, which consists of (a) reconstruction modules (left) and (b) animation modules (right). Given a source image $I_s$, we train an instant encoder $E$, adapted from \cite{lp3d}, to encode $I_s$ into a set of 3D Gaussians \cite{3dgs} for free viewpoint rendering (Sec.~\ref{sec:method-design}). %
For animation, we update the set of {Gaussians} {conditioned on an expression} from a driving image $I_d$, while preserving the appearance{ in $I_s$} (Sec.~\ref{sec:method-design}). We denote the encoding, animation and rendering procedure as $E(I_s, I_d, p)$ to synthesize a 2D image with identity from $I_s$ and expression from $I_d$ at viewpoint $p$.

For training, we construct a \textit{frontalized} synthetic extreme facial expression dataset (Sec.~\ref{sec:method-distill}), in which each identity is represented by multiple images with different expressions. We first frontalize the FFHQ~\cite{karras2019style} in-the-wild portraits and leverage a pre-trained diffusion-based portrait animation model~\cite{zhao2025x} to synthesize facial expressions on the frontalized views. Multi-view images are synthesized on the fly during the training by a pre-trained expert 2D-to-3D network~\cite{lp3d} for multi-view supervision. We then perform self-reenactment—using $I_d$ (the driving image) to drive $I_s$ (the source image) of the same identity—and optimize our model by minimizing the reconstruction loss between the reconstructed result and $I_s$, and the driven result and $I_d$ (Sec.~\ref{sec:method-distill}). %

During \textit{inference}, we directly input $I_s$ into our encoder $E$ once to reconstruct its 3D representation%
, and then animate the resulting set of {Gaussians} according to any given driving image $I_d$, whose identity may differ from that of $I_s$. %
Note that our animation pipeline, consisting of compact MLPs, does not require re-encoding the expensive 3D representation from $I_s$ and $I_d$ as in \cite{kirschstein2025avat3r,voodooxp,deng2024portrait4dv2}, leading to faster inference. %
The rendering of the {Gaussians} from arbitrary viewpoints is realized through rasterization \cite{3dgs}.

\subsection{Encoder Design and Animation Representation}
\label{sec:method-design}
\paragraph{3D Gaussians decoder.} LP3D \cite{lp3d} lifts single 2D image to a 3D avatar represented by a neural triplane representation. %
However, its implicit radiance field representation is less ideal for representing dynamics as opposed to 3D Gaussians. 3D Gaussians offer more independent control over each primitive and fast rendering speed while maintaining 3D consistency \cite{huang2024sc}.%

Therefore, we make a minimal change to adapt the architecture of \cite{lp3d} into using {3D Gaussians} while preserving its strong capability of faithfully lifting 2D images into 3D. We sample {3D Gaussians} from the encoded triplanes, as explored in \cite{barthel2024gaussian}. We will first explain how we decode {3D Gaussians} from {features sampled from neural triplane} and then clarify how we decide the sampling locations.

We first use $96$ channels for the encoded triplanes. 
Given a sampled 3D location $\mathbf{x}_i$, we project it onto each plane {and} sum %
{along the channel dimension to obtain} the feature vector ${f}_i$ using the first $48$ channels and retrieve a vector $\mathbf{m}_i$ using the remaining $48$ channels (see Fig.~\ref{fig:pipeline}). We call $\mathbf{m}_i$ a motion basis vector and explain its details later.

We replace the original NeRF-based decoder in \cite{lp3d} with an MLP $\Phi$ as shown in Fig.~\ref{fig:pipeline} with a single hidden layer of $96$ units and softplus activation functions to decode the feature vector into a set of attributes for a 3D Gaussian:
\begin{equation}
    \Phi({f}_i)=\{\mathbf{\mu}_i,\mathbf{s}_i,\mathbf{q}_i,o_i,\mathbf{c}_i\}.
\end{equation}
Therefore, we associate one 3D Gaussian with each sampled location and the aforementioned motion basis vector. %
The final image is synthesized using the differentiable renderer~\cite{3dgs} from the set of {3D Gaussians} and the specified camera parameters as shown in Fig.~\ref{fig:pipeline}. Notably, unlike previous works \cite{gagavatar,zhao2024invertavatar,deng2024portrait4dv2,voodooxp}, we do not use the 2D convolution refinement module to improve the rendered image's quality. 
We denote the rendered result of this encoded set of {Gaussians} at the viewpoint of $I_s$ as $I_s'$.

To decide where to sample {3D Gaussians} from the triplanes, different from \cite{barthel2024gaussian}, we do not have a paired pre-trained radiance field model to propose the sampling locations. Instead, we find that simply adapting the original ray shooting operation and two-pass importance sampling strategy in~\cite{mildenhall2020nerf,eg3d} already gives reasonable performance. %

Specifically, given a camera viewpoint, we shoot one ray for one pixel. We uniformly sample locations on each ray to decode 3D Gaussians, and then perform an additional importance sampling based on the opacity of previous decoded Gaussians to decode another set of Gaussians. Notice that the shooting resolution does not need to coincide with the rendering resolution. 
In practice, we use a sampling resolution of $64\times64$, but a resolution of $512\times512$ for rendering, which greatly improves efficiency. With $48$ sampled {Gaussians} on each ray, this configuration yields about $200K$ {Gaussians}, which is sufficient to render at $512\times512$ resolution \cite{kirschstein2024gghead,10.1145/3721238.3730737}.%

Although this sampling of {Gaussians} is inherently viewpoint-dependent, we find in practice that during inference, sampling from a fixed frontal viewpoint already produces a sufficiently dense and representative set of {Gaussians}. The same set can then be effectively reused for rendering from novel viewpoints, accelerating inference. 
During training, the {Gaussians} are instantiated from the final rendering viewpoint %
to ensure {view consistency. We explain this later.}

\paragraph{Feature-space deformation for animation.} Typically, dynamics with {Gaussians} are modeled by deforming the {Gaussian attributes directly}, i.e., updating their position, scaling and rotation vectors and optionally their color \cite{wu20244d,aligngaussians,liu2024dynamic,giebenhain2024npga}. However, we find that such a design has limited capacity {to model expressions} and {struggles with learning} the animation details in the training dataset. We hypothesize that it is because the learning of motion in the low-dimensional 3D space is more difficult compared to learning on a potentially smoother manifold in the high-dimensional feature vector space%
. We {thus} propose to deform the feature vector ${f}_i${ sampled from the triplanes}, which encodes information for \textit{all} Gaussian properties {and offers a richer deformation space}, based on motion signals.

\begin{figure}[t]
    \centering
    \includegraphics[width=0.8\linewidth]{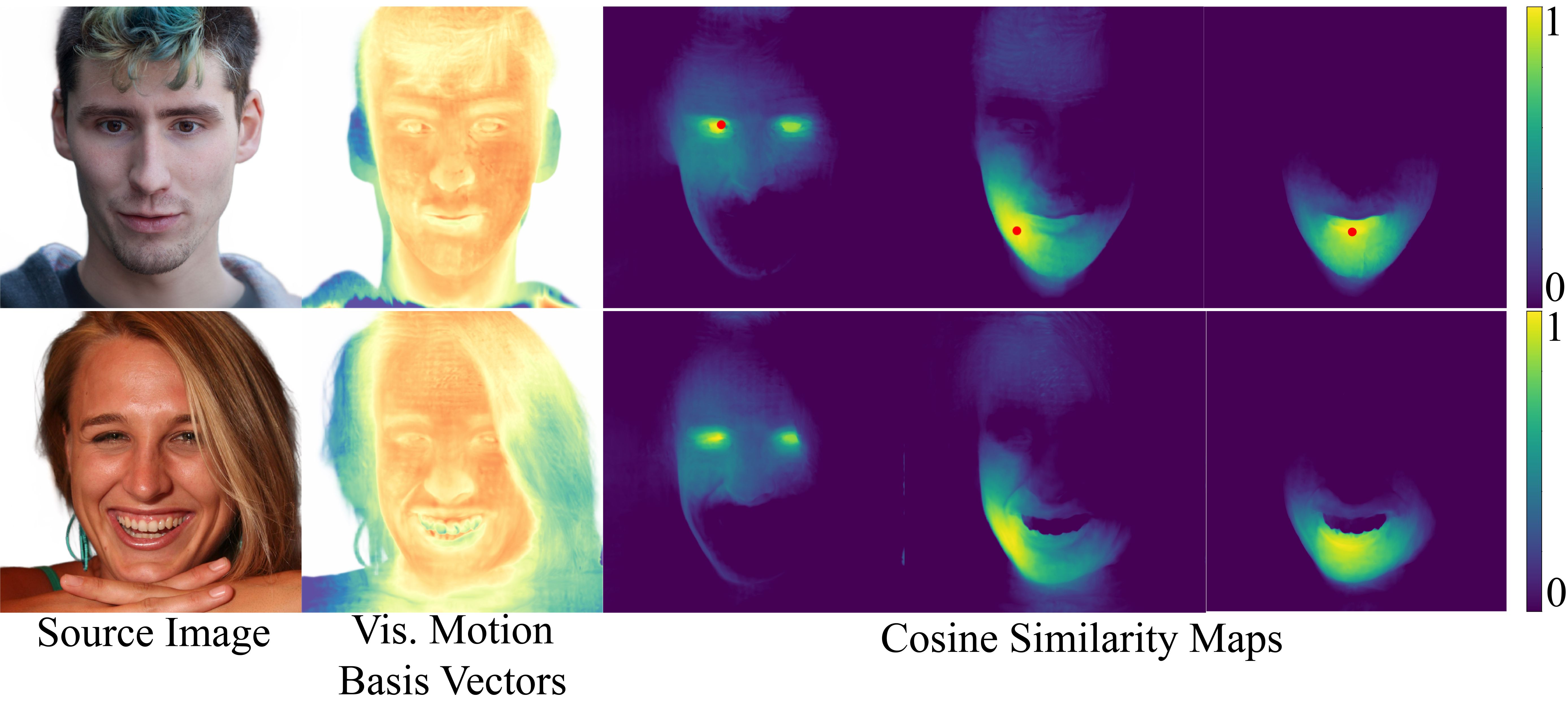}
    \caption{Demonstration of the similarity among motion basis vectors within and across subjects.
    Given the source images, we render the motion basis vectors of their Gaussian kernels via splatting.
    For the first-row subject, we select three specific points (red points) and compute the cosine similarity between their motion basis vectors and those of all other locations.
    We then compute the cosine similarity across subjects between these same points of the first subject and all motion basis vectors of the second subject in the second row.
    }
    \label{fig:face-correspondence}
\end{figure}
\begin{figure}[t]
    \centering
    \includegraphics[width=0.8\linewidth]{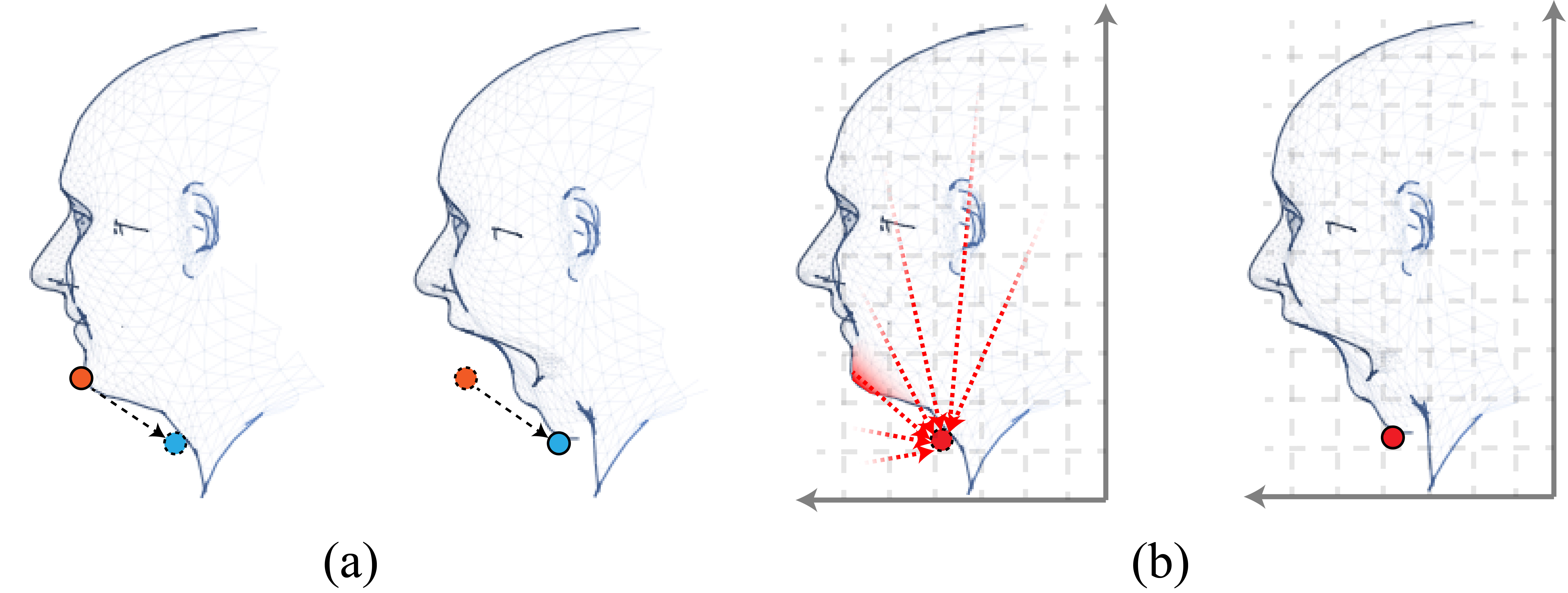}
    \caption{
    Conceptual comparison between predicting residual features per Gaussian versus per grid point on the triplanes \cite{deng2024portrait4dv2,voodooxp} in the case of realizing the expression of opening the mouth. %
    (a) In our framework, the 3D Gaussian can be transformed independently from the red point to the blue point because its motion basis vector encodes all necessary motion information. 
    (b) In contrast, existing triplanes-based works require aggregating dense global context, to update the features on each grid point.
    For example, it needs to fuse the shape information from the global context through the attention mechanism to decide whether the mouth will reach the red point and therefore update its geometry or not. %
    }
    \label{fig:residual_feature}
\end{figure}
\begin{figure*}[thb]
    \centering
    \includegraphics[width=0.95\linewidth]{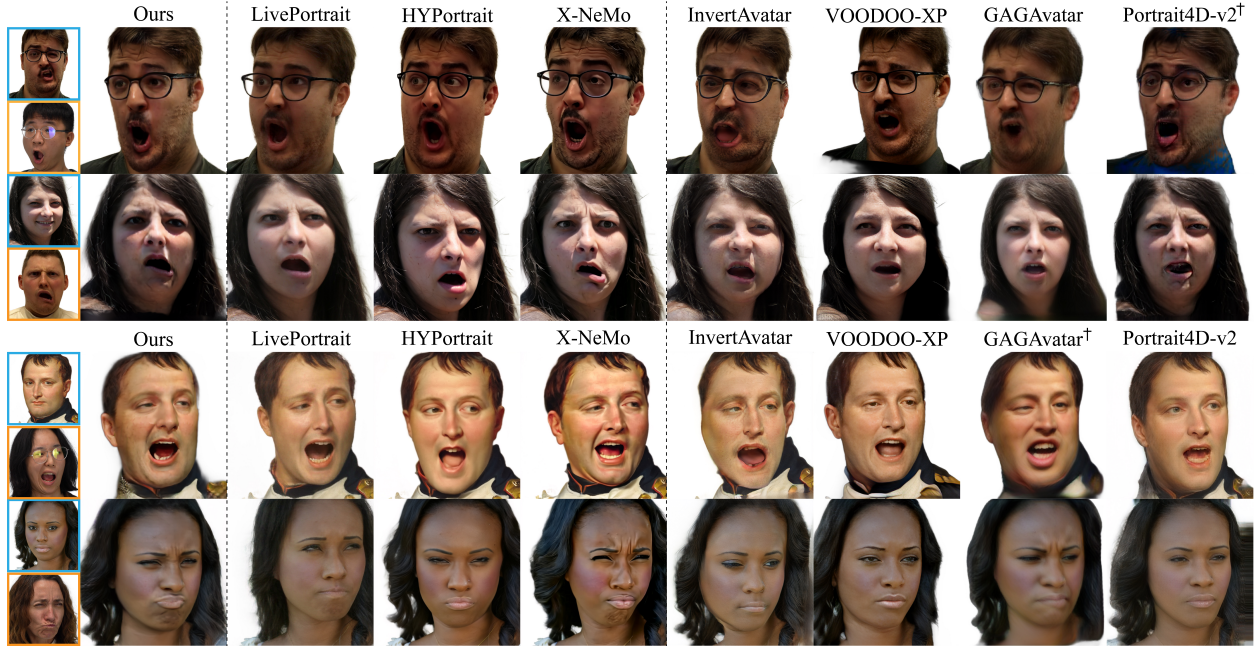}
    \caption{Qualitative comparison between our method and other 2D methods and 3D-aware methods in terms of expression and pose transfer. We denote \cite{xu2025hunyuanportrait} as ``HYPortrait''. \textcolor{srcblue}{\textbf{Source}} images are marked with blue borders at the leftmost column, while the \textcolor{drvorange}{\textbf{driving}} images are marked with orange borders throughout the paper. \textsuperscript{\textdagger} indicates that the methods are trained with our synthetic dataset for distillation from scratch for fair comparisons.%
    }
    \label{fig:qualitative}
\end{figure*}

Specifically, we first use the pre-trained motion encoder $M$ in \cite{zhao2025x} to encode the motion signals in the source image $I_s$ and the driving image $I_d$ into 1D motion coefficients $M(I_s), M(I_d)\in\mathbb{R}^{512}$, respectively (see Fig.~\ref{fig:pipeline}). Recall that, in Fig.~\ref{fig:pipeline}, each feature vector ${f}_i$ is associated with a motion basis vector $\mathbf{m}_i$, which encodes the spatially-varying deformation of the decoded 3D Gaussian and $\mathbf{m}_i$ does not change while animating. %
The concept is similar to facial muscles or localized bases in blendshape animation~\cite{Lewis2014BS}, 
but it is uniquely adapted for each individual and each facial part to capture personalized differences such as wrinkles. %
We show that the learned motion basis vector is coherent, semantically-meaningful and localized in Fig.~\ref{fig:face-correspondence}. 
Altogether, a residual feature $\delta f_i(I_s\rightarrow I_d)\in\mathbb{R}^{48}$ \textit{due to the motion} from $I_s$ to $I_d$ for each sampled Gaussian is \textit{individually} predicted by a motion decoder $\Psi$ {using} both the motion basis vector $\mathbf{m_i}$ %
and motion coefficients $M(I_s)$ and $M(I_d)$. %
Formally, the $i^\text{th}$ 3D Gaussian is updated as:
\begin{equation}
    \Phi({f}_i+\delta {f}_i(I_s\rightarrow I_d))=\{\mathbf{\mu}_i,\mathbf{s}_i,\mathbf{q}_i,o_i,\mathbf{c}_i\}, 
\end{equation}

\paragraph{Details of motion decoder $\Psi$.}
We adopt a single-layer AdaLN \cite{adaln} to modulate the motion basis vector $\mathbf{m}_i$ conditioned on the concatenated motion coefficients $M(I_s)$ and $M(I_d)$ and then pass the modulated vector to another MLP with a single hidden layer of $96$ units and softplus activation functions to predict the delta feature vector.%

\paragraph{Discussion.} We highlight the difference between our design of predicting residual features and the design of existing works (i.e., \cite{deng2024portrait4dv2,voodoo3d}) in Fig.~\ref{fig:residual_feature} with visual explanations.
Our design is more efficient by locally deforming each 3D Gaussian individually without aggregating dense global context, while also being expressive to capture motion details (Table~\ref{table:cross-reenactment}). %

\subsection{Training with Guidance from a Diffusion Model}
\label{sec:method-distill}
We adopt a state-of-the-art 2D facial animation diffusion model X-NeMo \cite{zhao2025x} as our data synthesizer. 
Specifically, we construct a synthetic dataset for self-reenactment-based training, containing over \num{60000} real identities from the FFHQ dataset \cite{karras2019style}, each of which has $8$ synthesized expressions with the driving expressions sampled from both the FFHQ and FEED datasets \cite{drobyshev2024emoportraits}. These two datasets altogether represent diverse identities and expressions in the real world. %
To minimize inconsistency and hallucination issues in the diffusion model, we use pre-trained LP3D \cite{lp3d,stengel2023} to frontalize the {source} and driving images before synthesizing expressions via X-NeMo. During the training, we apply the same LP3D on these frontalized synthetic portraits on the fly to randomize input and output viewpoints for augmentation and novel view supervision. %

\paragraph{Training procedure.} We use self-reenactment as the training objective. With the synthetic dataset described above, we first sample $I_s$ and $I_d$ from the same identity whose expressions could be same or different, and then randomly sample camera parameters as in \cite{lp3d} and estimate their {other} viewpoint images %
using a frozen pre-trained LP3D \cite{lp3d,stengel2023} for multi-view supervision, as shown in Fig.~\ref{fig:pipeline}. When $I_d$ has the same expression {as} $I_s$, we intend to learn {zero} residual features. 
Given the synthesized result $I'_d =E(I_s, I_d, p(I_d))$ and $I_d$, where $p(I_d)$ denotes the viewpoint of $I_d$ and Gaussians are therefore instantiated from $p(I_d)$, we design the loss {to compare} them along with an adversarial objective to enhance the image quality as:
\begin{equation}
\begin{aligned}
    \mathcal{L}=&\lambda_\text{L1}\mathcal{L}_\text{L1}(I'_d, I_d)+\lambda_\text{LPIPS}\mathcal{L}_\text{LPIPS}(I'_d, I_d)+\lambda_\text{ID}\mathcal{L}_\text{ID}(I'_d, I_d) + \\
    &\lambda_\text{Detail}\mathcal{L}_\text{Detail}(I'_d, I_d) + \lambda_\text{Norm}\mathcal{L}_\text{norm} + \lambda_\text{adv}\mathcal{L}_\text{adv}, 
\end{aligned}
\end{equation}
where $\lambda_\text{L1}=1$; $\mathcal{L}_\text{LPIPS}$ denotes the perceptual loss \cite{zhang2018perceptual} with $\lambda_\text{LPIPS}=1$; $\mathcal{L}_\text{ID}$ denotes the identity loss \cite{deng2019arcface} with $\lambda_\text{ID}=0.1$; $\lambda_\text{Detail}$ {computes an} additional separate L1 loss over the eye and mouth regions \cite{yu2021bisenet} with $\lambda_\text{Detail}=0.1$; $\lambda_\text{Norm}$ regularizes the averaged L2 norm for each predicted residual feature to enforce sparsity as in \cite{wu2022anifacegan,wang2025lia} with $\lambda_\text{Norm}=0.001$; $\mathcal{L}_\text{adv}$ is the adversarial loss as in \cite{lp3d}, which is further conditioned on the motion coefficients extracted by the motion encoder $M$, with $\lambda_\text{adv}=0.025$. Similarly, we also compute the loss for $I'_s$ and $I_s$ and sum it with loss for $I'_d$ and $I_d$ as our final loss. %

\paragraph{Implementation details.} We initialize all our networks including our instant lifting encoder and the adversarial discriminator used in $\mathcal{L}_\text{adv}$ from random weights, except the motion encoder $M$ from ~\cite{zhao2025x} that is pre-trained and frozen, and optimize with the Adam optimizer \cite{kingma2014adam} and batch size $32$ and learning rate $5e-5$%
. We gradually increase the rendering resolution from $64$ to $512$ with fixed number of sampled Gaussians and introduce the adversarial loss in the middle of training%
. More details are in the supplementary.

\begin{table}[t]
\centering
\small
\begin{tabular}{crccccccc}
\hline
\multicolumn{2}{c}{\textbf{Method}} & 
MEt3R$\downarrow$ & PSNR $\uparrow$ & SSIM$\uparrow$ & LPIPS$\downarrow$ & IoU $\uparrow$ & Mem.$\downarrow$ & FPS $\uparrow$ \\
\hline
\multicolumn{1}{l|}{\multirow{3}{*}{\rotatebox{90}{2D}}}
& \multicolumn{1}{r|}{LivePortrait \cite{guo2024liveportrait}} 
& 0.032 & 22.18 & 0.821 & 0.174 & 0.79 & \cellcolor{orange!25}0.5 GB & \cellcolor{orange!25}78.12 \\
\multicolumn{1}{l|}{}
& \multicolumn{1}{r|}{HYPortrait \cite{xu2025hunyuanportrait}} 
& 0.028 & 22.31 & {0.834} & \cellcolor{orange!25}{0.169} & \cellcolor{orange!25}{0.83} & 4.2 GB & 0.01 \\
\multicolumn{1}{l|}{}
& \multicolumn{1}{r|}{X-NeMo \cite{zhao2025x}} 
& 0.032 & 21.92 & 0.823 & 0.174 & \cellcolor{orange!25}{0.83} & $6.0$ GB & 0.03 \\
\hline
\multicolumn{1}{c|}{\multirow{7}{*}{\rotatebox{90}{3D}}}
& \multicolumn{1}{r|}{GAGAvatar \cite{gagavatar}} 
& 0.032 & 21.45 & 0.821 & 0.191 & 0.80 & 1.4 GB & 0.41 \\
\multicolumn{1}{c|}{}
& \multicolumn{1}{r|}{InvertAvatar \cite{zhao2024invertavatar}}  
& \cellcolor{orange!25}{0.026} & \cellcolor{red!25}{23.43} & \cellcolor{red!25}{0.846} & \cellcolor{red!25}{0.159} & 0.79 & 3.0 GB & 0.07 \\
\multicolumn{1}{c|}{}
& \multicolumn{1}{r|}{VOODOO-XP \cite{voodooxp}}    
& 0.030 & 21.14 & 0.809 & 0.197 & 0.77 & 1.2 GB & 5.45 \\
\multicolumn{1}{c|}{}
& \multicolumn{1}{r|}{Portrait4D-v2 \cite{deng2024portrait4dv2}} 
& 0.030 & 22.17 & 0.833 & 0.171 & 0.82 & 1.5 GB & 14.50 \\
\cline{2-9}
\multicolumn{1}{l|}{}
& \multicolumn{1}{r|}{GAGAvatar\textsuperscript{\textdagger} \cite{gagavatar}}  
& 0.027 & 21.33 & 0.826 & 0.199 & 0.77 & 1.4 GB & 0.41 \\
\multicolumn{1}{l|}{}
& \multicolumn{1}{r|}{Portrait4D-v2\textsuperscript{\textdagger} \cite{deng2024portrait4dv2}} 
& 0.030 & 21.25 & 0.822 & 0.195 & 0.82 & 1.5 GB & 14.50 \\
\multicolumn{1}{c|}{}
& \multicolumn{1}{r|}{Ours} 
& \cellcolor{red!25}{0.025} & \cellcolor{orange!25}{22.34} & \cellcolor{orange!25}{0.835} & 0.181 & \cellcolor{red!25}{0.85} & \cellcolor{red!25}$0.4$ GB & \cellcolor{red!25}107.31 \\
\hline
\end{tabular}
\caption{
Quantitative comparison on \textbf{self-reenactment}. 
Red indicates best and orange indicates second best throughout the paper. \textsuperscript{\textdagger} indicates training with our synthetic dataset.
}
\label{table:self-reenactment}
\end{table}

\begin{table}[t]
\centering
\small
\begin{tabular}{crcccccc}
\hline
\multicolumn{2}{c}{\textbf{Method}} & 
MEt3R$\downarrow$ & ID$\uparrow$ & EMO$\uparrow$ & AED$\downarrow$ & APD$\downarrow$ & FPS$\uparrow$ \\
\hline
\multicolumn{1}{l|}{\multirow{3}{*}{\rotatebox{90}{2D}}}
& \multicolumn{1}{r|}{LivePortrait \cite{guo2024liveportrait}} 
& 0.033 & 0.74 & 0.716 & 0.810 & \cellcolor{orange!25}{0.026} & \cellcolor{orange!25}{78.12} \\
\multicolumn{1}{l|}{}
& \multicolumn{1}{r|}{HYPortrait \cite{xu2025hunyuanportrait}} 
& 0.032 & 0.74 & 0.752 & 0.900 & 0.077 & 0.01 \\
\multicolumn{1}{l|}{}
& \multicolumn{1}{r|}{X-NeMo \cite{zhao2025x}} 
& 0.035 & 0.72 & \cellcolor{orange!25}{0.760} & \cellcolor{orange!25}{0.805} & 0.032 & 0.03 \\
\hline
\multicolumn{1}{c|}{\multirow{7}{*}{\rotatebox{90}{3D}}}
& \multicolumn{1}{r|}{GAGAvatar \cite{gagavatar}} 
& 0.034 & 0.77 & 0.654 & 0.888 & \cellcolor{red!25}{0.025} & 0.41 \\
\multicolumn{1}{c|}{}
& \multicolumn{1}{r|}{InvertAvatar \cite{zhao2024invertavatar}}  
& \cellcolor{red!25}{0.028} & \cellcolor{red!25}{0.80} & 0.565 & 0.891 & 0.049 & 0.07 \\
\multicolumn{1}{c|}{}
& \multicolumn{1}{r|}{VOODOO-XP \cite{voodooxp}}    
& 0.032 & 0.77 & 0.699 & 0.903 & 0.028 & 5.45 \\
\multicolumn{1}{c|}{}
& \multicolumn{1}{r|}{Portrait4D-v2 \cite{deng2024portrait4dv2}} 
& 0.035 & \cellcolor{orange!25}{0.79} & 0.589 & 0.886 & 0.029 & 14.50 \\
\cline{2-8}
\multicolumn{1}{l|}{}
& \multicolumn{1}{r|}{GAGAvatar\textsuperscript{\textdagger} \cite{gagavatar}}  
& \cellcolor{orange!25}{0.030} & 0.67 & 0.578 & 0.944 & 0.038 & 0.41 \\
\multicolumn{1}{l|}{}
& \multicolumn{1}{r|}{Portrait4D-v2\textsuperscript{\textdagger} \cite{deng2024portrait4dv2}} 
& 0.034 & 0.76 & 0.448 & 0.972 & \cellcolor{orange!25}{0.026} & 14.50 \\
\multicolumn{1}{c|}{}
& \multicolumn{1}{r|}{Ours} 
& \cellcolor{red!25}{0.028} & 0.76 & \cellcolor{red!25}{0.785} & \cellcolor{red!25}{0.731} & 0.028 & \cellcolor{red!25}{107.31} \\
\hline
\end{tabular}
\caption{
Quantitative comparison on \textbf{cross-reenactment}.
}
\label{table:cross-reenactment}
\end{table}

\section{Results}
\label{sec:results}
We provide quantitative and qualitative comparisons and an ablation study here. For more results, including a real-time demo, please refer to the supplementary material and the accompanying video.
\paragraph{Metrics.} We conduct experiments for facial animation using the VOODOO-XP test set as in \cite{voodooxp}, which contains $102$ video sequences. It features extreme expressions and wide viewing angles. We extract one out of five consecutive frames for testing to eliminate unnecessary duplication, which, in total, results in over $20K$ images. We evaluate results on common head regions without the background across all methods. 
We conduct the following experiments. (1) Self-reenactment: for each video sequence, we use the first frame as the source frame, and all other frames to drive it. Each method is tasked with reproducing the viewpoint and the expression of the driving frame. (2) Cross-reenactment: for each video sequence, we randomly sample another video sequence. We use the first frame of the first video sequence as the source frame, and all frames in the second video sequence to drive it.%

Besides the memory consumption for static model storage required during inference and the rendering speed measured in FPS on an NVIDIA 6000 Ada GPU, we evaluate performance using the following five aspects:
(a) MEt3R~\cite{asim24met3r} for dense 3D inconsistency; (b) PSNR, SSIM~\cite{ssim} and LPIPS~\cite{zhang2018perceptual} to evaluate the quality of image reconstruction; (c) IoU to evaluate 3D reconstruction quality by computing the intersection-over-union between the binary silhouettes of the full head region in the driving frame and the resulting frame of the same subject; (d) face ID consistency \cite{deng2019arcface}; (e) the accuracy of expression and pose transfer where we use SMIRK~\cite{smirk} to extract the FLAME~\cite{FLAME:SiggraphAsia2017} coefficients for the driving and resulting frames, and measure the averaged distance of the expression coefficients as ``AED'', and averaged distance of the pose {parameters} as ``APD''. Besides, we also use EmoNet~\cite{emonet} as in \cite{zhao2025x} to measure the emotion similarity between the driving and resulting frames as ``EMO'' that is more sensitive to extreme motions. 
Please note that ``APD'' reveals \textit{coarse} 3D shape quality while MEt3R focuses more on dense photometric 3D consistency across views.

\subsection{Comparisons}
\paragraph{Baselines.} We compare our method against other{ existing} open-source feed-forward methods, %
including the state-of-the-art GAN-based 2D facial animation method LivePortrait \cite{guo2024liveportrait}, diffusion-based 2D facial animation methods HunyuanPortrait~\cite{xu2025hunyuanportrait} and X-NeMo \cite{zhao2025x}, 3DMM-based 3D facial animation methods InvertAvatar~\cite{zhao2024invertavatar} and GAGAvatar \cite{gagavatar}, and 3D facial animation methods with learned motion space Portrait4D-v2~\cite{deng2024portrait4dv2} and VOODOO-XP \cite{voodooxp}. Furthermore, {for an additional fairer comparison and} to illustrate the benefits of our proposed animation representation, we also train {the best-performing prior 3D-based methods} Portrait4D-v2 and GAGAavatar from scratch using our synthetic dataset with multi-view images estimated and denote them as Portrait4D-v2\textsuperscript{\textdagger} and GAGAvatar\textsuperscript{\textdagger}.

\begin{figure}[t]
    \centering
    \includegraphics[width=\linewidth]{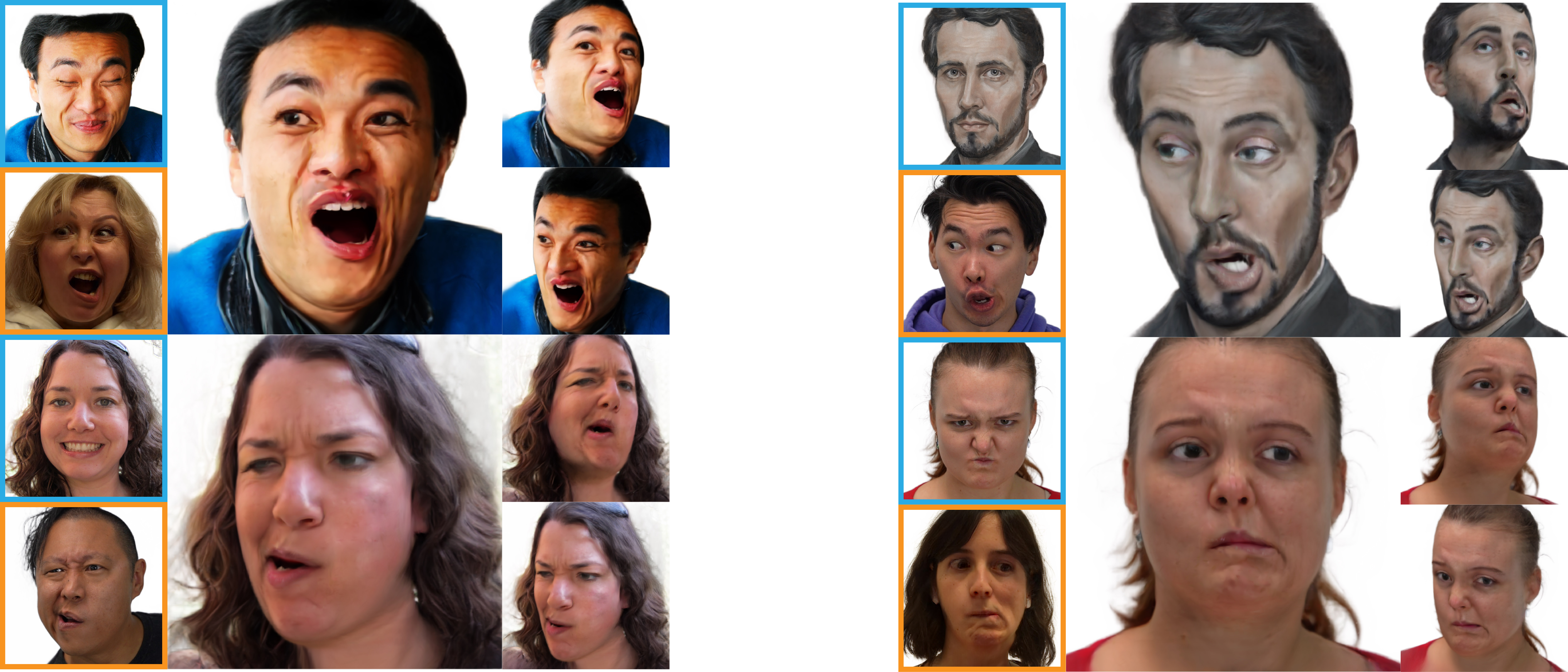}
    \caption{More qualitative results with our method from a \textcolor{srcblue}{\textbf{source}} image and \textcolor{drvorange}{\textbf{driving}} image. We provide the multi-view rendered results next to the driven results.%
    }
    \label{fig:more-results}
\end{figure}

\paragraph{Qualitative results.} We provide qualitative comparisons in Fig.~\ref{fig:qualitative}.
Generally, other 3D-aware methods create muted expressions even when retrained with our synthetic dataset and do not faithfully synthesize expressions in the driving image. Especially, in the first row, the 3D-aware methods cannot remove the extreme mouth motion in the source frame and the right wrinkles near the mouth are leaked into the driven results. InvertAvatar occasionally produces collapsed 3D head geometry (third and forth rows). Among the 2D-based methods, X-NeMo mostly produces impressive results, but occasionally distorts head shape under a large pose change (third row) and hallucinates details (e.g., added cheek color in the fourth row). HunYuanPortrait cannot faithfully transfer the pose and expression in the second, third and fourth rows. 
LivePortrait creates dampened mouth expressions in the second to fourth rows.
Even after retraining with our expressive synthetic dataset, Portrait4D-v2\textsuperscript{\textdagger} and GAGAvatar\textsuperscript{\textdagger} produce less accurate expression transfer results.
In contrast, our method faithfully transfers the expression and pose and is on par with X-NeMo while maintaining the identity and being $3500\times$ faster (see Tab.~\ref{table:cross-reenactment}). Please find more comparison in the supplementary.
We further provide multi-view rendering results in Fig.~\ref{fig:more-results}. Despite extreme expressions being present in the source images, our method can infer the occluded regions, such as the closed eyes, produces outputs with identity consistent with that of the source image and motions consistent with the driving image. 

\paragraph{Quantitative results.} 
As shown in Table~\ref{table:self-reenactment},~\ref{table:cross-reenactment}, our method achieves the best 3D consistency (MEt3R), 3D shape (IOU), and compute efficiencies (Mem and FPS) and second best image reconstruction (in terms of PSNR and SSIM) across all methods, for the task of self-reenactment. All of the compared 3D baselines utilize camera-space 2D refinement to improve image quality, but this is known to degrade 3D consistency -- reflected by their lower MEt3R score. Even though InvertAvatar achieves the best image quality, its 3D quality is significantly limited, as captured by the worse IoU for the 3D head shape compared to us.
We also achieve state-of-the-art MEt3R, EMO, and AED scores across all methods for the task of cross-reenactment, even surpassing X-NeMo due to potentially more accurate pose transfer, which aids correct expression and emotion detection. We note that ID scores conflate identity preservation with expression transfer accuracy, since simply keeping the source image without modifying the expression can maximize the ID score (please see more analysis in the supplement). For example, InvertAvatar achieves strong identity similarity (high ID) but fails to faithfully reproduce the target expression in Fig.~\ref{fig:qualitative} (fourth row), resulting in worse EMO and AED scores in Table~\ref{table:cross-reenactment}. InvertAvatar's collapsed 3D shape is also reflected in poor APD due to inaccurate pose transfer. 

While our training data is synthesized using X-NeMo and LP3D, we restrict X-NeMo to operate on frontal images only and use LP3D to estimate multi-view images to resolve identity drift. This improves consistency beyond that of the teacher model (X-NeMo) as evidenced by our method's better ID metric and consistency metric versus X-NeMo (Table~\ref{table:cross-reenactment}).%

Our method animates faces at 107.31 FPS, surpassing all other baselines, including ones using attention mechanisms \cite{deng2024portrait4dv2,voodooxp}. In contrast, 2D diffusion methods could take up to almost a minute for driving a single image sequentially, or several seconds per frame when a sequence of images is used and the time cost is amortized across frames. For our method, encoding a facial image into 3D Gaussians is required only once {per video sequence}, and our method performs this step almost instantly in just $20$ms.

\begin{table}[t]
\centering
\small
\begin{tabular}{l|cccccccc}
\hline
\multicolumn{1}{c|}{\textbf{Method}} & MEt3R $\downarrow$ & PSNR $\uparrow$ & SSIM $\uparrow$ & LPIPS $\downarrow$ & {AED}$\downarrow$ & IoU$\uparrow$ & {Mem.}$\downarrow$ & {FPS}$\uparrow$ \\ \hline
Ours ($128\times 128$) & \cellcolor{red!25}0.073 & \cellcolor{orange!25}23.18 & \cellcolor{orange!25}0.785 & \cellcolor{red!25}0.104 & \cellcolor{red!25}{0.507} & \cellcolor{orange!25}0.88 & \cellcolor{orange!25}{0.32 GB} & \cellcolor{orange!25}{132.87} \\
\quad {w/ \scriptsize DINO-v2 Encoding} & \cellcolor{orange!25}0.074 & 22.88 & 0.776 & 0.110 & 0.597 & 0.87 & 4.32 GB & 22.47 \\
\quad {w/ \scriptsize Real Dataset} & \cellcolor{red!25}0.073 & 23.10 & 0.782 & \cellcolor{orange!25}0.109 & \cellcolor{orange!25}{0.543} & \cellcolor{orange!25}0.88 & \cellcolor{orange!25}{0.32 GB} & \cellcolor{orange!25}{132.87} \\
\quad {w/ \scriptsize Spatial Deformation} & 0.075 & \cellcolor{red!25}23.27 & \cellcolor{red!25}0.788 & \cellcolor{orange!25}0.109 & 0.634 & \cellcolor{red!25}0.89 & \cellcolor{red!25}{0.27 GB} & \cellcolor{red!25}{141.61} \\
\hline
\end{tabular}
\caption{Ablation studies with self-reenactment.
}
\label{table:ablation}
\end{table}

\subsection{Ablation Studies}
\label{sec:ablation}
\begin{figure}[t]
    \centering
    \includegraphics[width=0.8\linewidth]{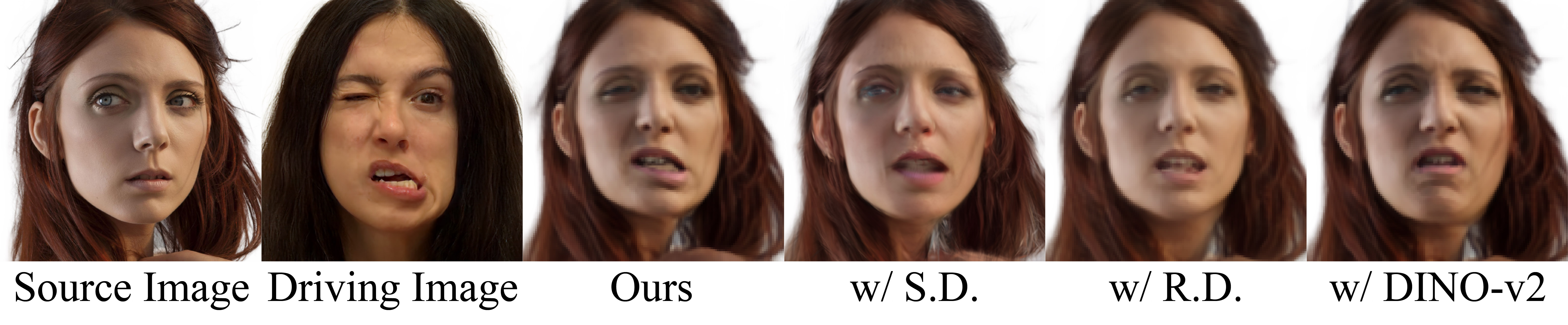}
    \caption{Comparison among different ablation models based on the expression transfer. ``w/ S.D.'' denotes using the spatial deformation instead of feature-space deformation. ``w/ R.D.'' denotes using a real dataset instead of the synthetic one.}
    \label{fig:ablation}
\end{figure}

We validate each component of our model using the self-reenactment task on the same VOODOO-XP test set with a {rendering} resolution of $128\times128$ without an adversarial loss for more stable comparison. %
We use the exact same metrics for quantitatively evaluating self-reenactment as before along with AED for measuring expression transfer accuracy, reported in Table~\ref{table:ablation} and Fig.~\ref{fig:ablation}. %

\paragraph{Choice of motion encoder.} We study the importance of the selected motion encoder. Instead of the motion encoder in X-Nemo \cite{zhao2025x} we use DINO-v2~\cite{oquab2023dinov2} as in \cite{voodooxp}. Notably, this encoder is computationally more expensive than our selected motion encoder \cite{zhao2025x}. It increases the memory consumption and reduces the FPS. It fails to capture expression details in Fig.~\ref{fig:ablation} and leads to a worse AED.

\paragraph{Usage of diffusion model.} We study the necessity of using a synthetic dataset generated with a pre-trained diffusion model by using a real dataset instead. %
We use CelebVText \cite{yu2022celebvtext} as the real dataset for its diversity of expressions and identities. Since a real dataset usually features common expressions such as smiling, opening the mouth, etc. and less extreme expressions, the AED is affected and the produced expression is muted (Fig.~\ref{fig:ablation}).

\paragraph{Feature-space- vs. spatial-deformation.} %
We ablate the usage of the proposed feature-space deformation by replacing it with directly predicting the residual for the 3D Gaussians' position, scaling and quaternion vectors from the motion decoder $\Psi$, similarly to the previous dynamic modeling methods (e.g., \cite{wu20244d,aligngaussians,liu2024dynamic,giebenhain2024npga}). Since we do not need to go through the non-linear decoder in this setup, the memory consumption and FPS are slightly improved. However, the expression transfer accuracy is significantly compromised, leading to worsened AED. This is also reflected in Fig.~\ref{fig:ablation}.

\section{Discussion}
\label{sec:conclusion}

\paragraph{Conclusion.} In this work, we propose an instant 4D method that encodes a single image into 3D-consistent, efficient yet expressive avatar.
We propose an animation representation that deforms both the Gaussian appearance and geometry based on the encoded motion basis vectors, inspired by bases representations in traditional facial animation. %
We believe our method paves the way for a {real-time} expressive representation%
, enabling real-world applications such as digital twins, where real-time performance, and controllability are critical. 

\paragraph{{Limitation \& }Future work.} %
{Our method finds it hard to extend to high-quality eyewear reconstruction due to thin structures, and complex reflection and refraction.}
We plan to extend our method into disentangling appearance properties (lighting). %
Besides, even though we demonstrate our method using only image-driven examples, it is possible to encode other conditions, such as audio or texts, into sequences of the 1D motion coefficients and drive our avatar.

\paragraph{Ethics concern.} We propose an algorithm, which is capable of converting a single 2D facial image into a 3D-aware animatable avatar, which could be misused for generating malicious content. 
We do not condone such behavior and identify potential works that could be used for detecting fake information~\cite{deep-fake, DeepfakeBench_YAN_NEURIPS2023, UCF_YAN_ICCV2023, LSDA_YAN_CVPR2024} or works that authenticate the authorized driving subject of the avatar~\cite{prashnani2024avatar}. %

\section*{Acknowledgements}
We thank David Luebke, Michael Stengel, Yeongho Seol, Simon Yuen, Marcel B\"uhler, and Arash Vahdat for feedback on drafts and early discussions. We also thank Alex Trevithick and Tianye Li for proof-reading. This research was also funded in part by the Ronald L. Graham Chair and the UC San Diego Center for Visual Computing.

\bibliographystyle{splncs04}
\bibliography{main}

\end{document}